\documentclass[UTF8]{article} 
\usepackage{iclr2020_conference,times}

\usepackage{amsmath,amsfonts,bm}

\def\eqref#1{equation~\ref{#1}}

\def\1{\bm{1}}

\DeclareMathAlphabet{\mathsfit}{\encodingdefault}{\sfdefault}{m}{sl}
\SetMathAlphabet{\mathsfit}{bold}{\encodingdefault}{\sfdefault}{bx}{n}

\usepackage{hyperref}
\usepackage{url}
\usepackage{booktabs}       
\usepackage{amsfonts}       
\usepackage{nicefrac}       
\usepackage{microtype}      
\usepackage{lipsum}
\usepackage[font=small,labelfont=bf]{caption} 

\usepackage{graphicx}
\usepackage{grffile}
\graphicspath{ {/Images/} }
\title{Amharic Abstractive Text Summarization}

\author{Amr M.~Zaki \thanks{(https://medium.com/@theamrzaki/) (https://github.com/theamrzaki/ )} \\
	Department of Computer Engineering\\
	Ain-Shams University\\
	\texttt{theamrzaki@hotmail.com} 
	\And
	Mahmoud I. Khalil \\
	Department of Computer Engineering \\
	Ain-Shams University \\
	\texttt{mahmoud.khalil@eng.asu.edu.eg} \\
	\AND
	Hazem M. Abbas \\
	Department of Computer Engineering \\
	Ain-Shams University \\
	\texttt{hazem.abbas@eng.asu.edu.eg}
}

\iclrfinalcopy 
\begin{document}

\maketitle

\begin{abstract}
Text Summarization is the task of condensing long text into just a handful of sentences. Many approaches have been proposed for this task, some of the very first were building statistical models (Extractive Methods [\cite{paice1990constructing}],[\cite{kupiec1999trainable}]) capable of selecting important words and copying them to the output, however these models lacked the ability to paraphrase sentences, as they simply select important words without actually understanding their contexts nor understanding their meaning, here comes the use of Deep Learning based architectures (Abstractive Methods [\cite{chopra-etal-2016-abstractive}],[\cite{seq2seqANDbeyond}]), which effectively tries to understand the meaning of sentences to build meaningful summaries. In this work we discuss one of these new novel approaches which combines curriculum learning with Deep Learning, this model is called \textbf{Scheduled Sampling} [\cite{scheduledsamplingCL}]. We apply this work to one of the most widely spoken African languages which is the Amharic Language, as we try to enrich the African NLP community with top-notch Deep Learning architectures.
\end{abstract}

\section{Dataset and Word-Embedding for the Amharic Language}
Working with the Amharic Language turned to be quite challenging, as African Languages are typically known to be low resource languages. Data for our task wasn't easily collected, as there isn't an available dataset for our task, this is why we had to collect and build our own dataset from scratch. Data for text summarization is found in form of long text (articles) and their summaries (titles), for the English case, researchers work on data scrapped from CNN/DailyNews [\cite{cnndailyDataset}], so we used their same approach and scrapped data from 7 well known Amharic News websites,
\begin{enumerate}
	\item \url{http://www.goolgule.com/}
	\item \url{https://www.ethiopianregistrar.com/amharic}
	\item \url{https://amharic.ethsat.com} 
	\item \url{https://ecadforum.com/Amharic}
	\item \url{https://www.zehabesha.com/amharic/}
	\item \url{https://www.ethiopianregistrar.com/amharic/}
	\item \url{https://www.ethiopianreporter.com/article/}
\end{enumerate} we scrapped over 50k articles, and only used those with long titles (about 19k articles).\\
Word-embedding has proved itself as one of the best methods to represent text for deep-models. One of the most widely used English word-embedding models is Word2Vec[\cite{word2vec}], it represents each word with a list of vectors to be easily used in the deep-models, however no such models were trained for the Amharic Language, this is why we trained our own model for this task.\\
In this work, we provide both the scrapped news dataset, and the trained word-embedding as open source to help enrich the African NLP research community \footnote{https://github.com/theamrzaki/text\_summurization\_abstractive\_methods/tree/master/Amharic}.

\section{Text Summarization Deep Learning Building Blocks}
Text Summarization is considered as a time-series problem, as we are trying to generate the next word, given the past words. Novel deep models rely on some basic blocks, in this section we go through these building blocks.

\subsection{Seq2Seq using LSTM with attention}
Since our task is a time-series problem, RNN models were first used to address this task, however given the long sentence dependencies in natural languages, LSTM based architectures were used given its memory structure [\cite{LSTM}]. Our task can actually be seen as mapping between input and output, however since they differ in length (long input, short output), seq2seq based architectures are used [\cite{seq2seqANDbeyond}]. To give our models even more human like abilities in summarization, [\cite{bahdanau2014neural}] suggested building a deep model on top of the seq2seq architecture, which helped it attend to important words in the input. 

\subsection{Pointer Generator Model}
This previously discussed model has a well-known problem, which is working with unknown out-of-vocab words, as it can only be trained on a fixed sized vocabulary. A solution was proposed by [\cite{seq2seqANDbeyond}],[\cite{get2point}] which builds a deep model on top of the seq2seq architecture capable of learning when to copy words, and when to generate new ones.

\section{Scheduled Sampling}
One of the problems that the above seq2seq based architecture suffers from, comes from the way it is trained, as the model is trained by supplying it both an input long text, and a reference short summary, while when we test the model, we only supply it with the input long text, and no reference is given. This forms an inconsistency between the training phase and the testing phase, as the model has never been trained to depend on its own, this problem is called \textbf{Exposure Bias} [\cite{ExposureBias}].\\
A solution proposed by [\cite{scheduledsamplingCL}] helped in solving this problem, which included combining curriculum learning with our deep model. We start the training normally by supplying both the long training text and the reference summary, but when the model becomes mature enough, we gradually introduce the model to its own mistakes while training, decreasing its dependency on the reference sentence, teaching the model to depend on itself in the training phase, in other words making the learning problem more difficult while the model matures, hence curriculum learning. 

\begin{minipage}{1\linewidth}		
	\centering
	\includegraphics[width=0.9\linewidth]{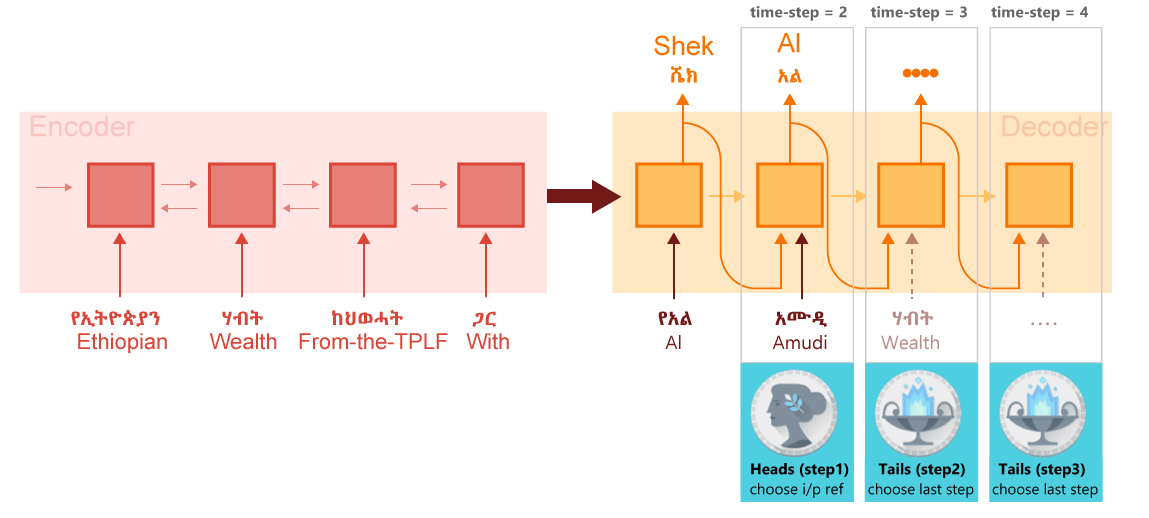}
	\captionof{figure}{Scheduled Sampling Architecture}
	\label{Scheduled Sampling Architecture}
\end{minipage}

\section{Experiments}
We have applied the \textbf{Scheduled Sampling} model on the Amharic Dataset that we have built, we have used google colab as our training framework, as it provides us with free GPU and up to 24GB of RAM. Our model is built over the library [\cite{DeepRLmethods}], we have modified it to work on python3 and to work with the Amharic Dataset.
We evaluate our experiments using well-known metrics used for evaluating text summarization, these metrics are ROUGE[\cite{Rouge}] and BLEU[\cite{Bleu}], which measure the amount of n-grams that overlap between the reference summary and our generated one, as the measure increases the amount of overlap increases, indicating a better output.

\begin{minipage}{1\linewidth}		
	\centering
	\includegraphics[width=1\linewidth]{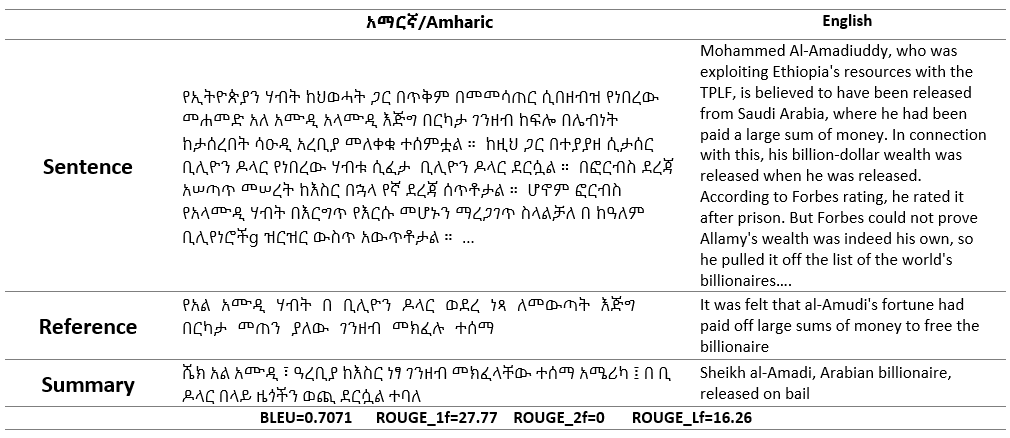}
\end{minipage}

We ran our evaluation on 100 test sentences, scores were \textbf{BELU=0.3311} , \textbf{ROUGE\_1f=20.51} , \textbf{ROUGE\_2f=08.59} , \textbf{ROUGE\_Lf=14.76}.

For comparison, running scheduled sampling on English well-known datasets (CNN/DailyMail dataset) achieves ROUGE-1 of \textbf{39.53} and \textbf{17.28} ROUGE-2, this discrepancy of the results from the English counterpart, comes from the fact that the English dataset is huge (200k articles with long summary) compared to our scrapped Amharic dataset of (19k articles with short summaries), this comes from the fact that collecting English dataset is comparatively much easier than collecting an African one, due to the huge amount of available English resources.

\section{Conclusion}
By building a custom word embedding model for a specific African language, we are able to apply any deep model (that works on English) on that selected African Language, like what we have proven by our work.

In our coming work, we are willing to experience with other advanced architectures that have recently proven extremely efficient in addressing seq2seq problems. One of these architectures is BERT [\cite{BERT}], which stands for Bidirectional Encoder Representations from Transformers, it uses a similar encoder-decoder architecture, but instead of using recurrent based cells, it only uses attention (self-attention). One efficient way to use BERT, is by using an already pre-trained model, that has been trained on the English dataset (CNN/DailyMail dataset), and then apply cross-lingual transfer to the Amharic dataset, we believe that by this, we believe that this may actually result in better summaries in spite of the relatively small Amharic dataset.

We hope that by this work, we have helped pave the way in applying novel Deep Learning techniques for African Languages, we also hope that we have contributed a guideline in applying deep models that can be further used in other NLP tasks.
\pagebreak

\bibliography{iclr2020_conference}

\begin{thebibliography}{15}
\providecommand{\natexlab}[1]{#1}
\providecommand{\url}[1]{\texttt{#1}}
\expandafter\ifx\csname urlstyle\endcsname\relax
  \providecommand{\doi}[1]{doi: #1}\else
  \providecommand{\doi}{doi: \begingroup \urlstyle{rm}\Url}\fi

\bibitem[Bahdanau et~al.(2014)Bahdanau, Cho, and Bengio]{bahdanau2014neural}
Dzmitry Bahdanau, Kyunghyun Cho, and Yoshua Bengio.
\newblock Neural machine translation by jointly learning to align and
  translate.
\newblock \emph{arXiv preprint arXiv:1409.0473}, 2014.

\bibitem[Bengio et~al.(2015)Bengio, Vinyals, Jaitly, and
  Shazeer]{scheduledsamplingCL}
Samy Bengio, Oriol Vinyals, Navdeep Jaitly, and Noam Shazeer.
\newblock Scheduled sampling for sequence prediction with recurrent neural
  networks.
\newblock In \emph{Proceedings of the 28th International Conference on Neural
  Information Processing Systems - Volume 1}, NIPS’15, pp.\  1171–1179,
  Cambridge, MA, USA, 2015. MIT Press.

\bibitem[Chopra et~al.(2016)Chopra, Auli, and
  Rush]{chopra-etal-2016-abstractive}
Sumit Chopra, Michael Auli, and Alexander~M. Rush.
\newblock Abstractive sentence summarization with attentive recurrent neural
  networks.
\newblock In \emph{Proceedings of the 2016 Conference of the North {A}merican
  Chapter of the Association for Computational Linguistics: Human Language
  Technologies}, pp.\  93--98, San Diego, California, June 2016. Association
  for Computational Linguistics.
\newblock \doi{10.18653/v1/N16-1012}.
\newblock URL \url{https://www.aclweb.org/anthology/N16-1012}.

\bibitem[Devlin et~al.(2019)Devlin, Chang, Lee, and Toutanova]{BERT}
Jacob Devlin, Ming-Wei Chang, Kenton Lee, and Kristina Toutanova.
\newblock {BERT}: Pre-training of deep bidirectional transformers for language
  understanding.
\newblock In \emph{Proceedings of the 2019 Conference of the North {A}merican
  Chapter of the Association for Computational Linguistics: Human Language
  Technologies, Volume 1 (Long and Short Papers)}, pp.\  4171--4186,
  Minneapolis, Minnesota, June 2019. Association for Computational Linguistics.
\newblock \doi{10.18653/v1/N19-1423}.
\newblock URL \url{https://www.aclweb.org/anthology/N19-1423}.

\bibitem[Hermann et~al.(2015)Hermann, Kocisky, Grefenstette, Espeholt, Kay,
  Suleyman, and Blunsom]{cnndailyDataset}
Karl~Moritz Hermann, Tomas Kocisky, Edward Grefenstette, Lasse Espeholt, Will
  Kay, Mustafa Suleyman, and Phil Blunsom.
\newblock Teaching machines to read and comprehend.
\newblock In C.~Cortes, N.~D. Lawrence, D.~D. Lee, M.~Sugiyama, and R.~Garnett
  (eds.), \emph{Advances in Neural Information Processing Systems 28}, pp.\
  1693--1701. Curran Associates, Inc., 2015.
\newblock URL
  \url{http://papers.nips.cc/paper/5945-teaching-machines-to-read-and-comprehend.pdf}.

\bibitem[Hochreiter \& Schmidhuber(1997)Hochreiter and Schmidhuber]{LSTM}
Sepp Hochreiter and J\"{u}rgen Schmidhuber.
\newblock Long short-term memory.
\newblock \emph{Neural Comput.}, 9\penalty0 (8):\penalty0 1735–1780, November
  1997.
\newblock ISSN 0899-7667.
\newblock \doi{10.1162/neco.1997.9.8.1735}.
\newblock URL \url{https://doi.org/10.1162/neco.1997.9.8.1735}.

\bibitem[Keneshloo et~al.(2019)Keneshloo, Shi, Ramakrishnan, and
  Reddy]{DeepRLmethods}
Yaser Keneshloo, Tian Shi, Naren Ramakrishnan, and Chandan Reddy.
\newblock Deep reinforcement learning for sequence-to-sequence models.
\newblock \emph{IEEE Transactions on Neural Networks and Learning Systems},
  PP:\penalty0 1--21, 08 2019.
\newblock \doi{10.1109/TNNLS.2019.2929141}.

\bibitem[Kupiec et~al.(1995)Kupiec, Pedersen, and Chen]{kupiec1999trainable}
Julian Kupiec, Jan~O. Pedersen, and Francine Chen.
\newblock A trainable document summarizer.
\newblock In \emph{SIGIR}, 1995.

\bibitem[Lin(2004)]{Rouge}
Chin-Yew Lin.
\newblock {ROUGE}: A package for automatic evaluation of summaries.
\newblock In \emph{Text Summarization Branches Out}, pp.\  74--81, Barcelona,
  Spain, July 2004. Association for Computational Linguistics.
\newblock URL \url{https://www.aclweb.org/anthology/W04-1013}.

\bibitem[Mikolov et~al.(2013)Mikolov, Corrado, Chen, and Dean]{word2vec}
Tomas Mikolov, G.s Corrado, Kai Chen, and Jeffrey Dean.
\newblock Efficient estimation of word representations in vector space.
\newblock pp.\  1--12, 01 2013.

\bibitem[Nallapati et~al.(2016)Nallapati, Zhou, dos Santos,
  GuÌ‡l{\c{c}}ehre, and Xiang]{seq2seqANDbeyond}
Ramesh Nallapati, Bowen Zhou, Cicero dos Santos, {\c{C}}a{\u{g}}lar
  GuÌ‡l{\c{c}}ehre, and Bing Xiang.
\newblock Abstractive text summarization using sequence-to-sequence {RNN}s and
  beyond.
\newblock In \emph{Proceedings of The 20th {SIGNLL} Conference on Computational
  Natural Language Learning}, pp.\  280--290, Berlin, Germany, August 2016.
  Association for Computational Linguistics.
\newblock \doi{10.18653/v1/K16-1028}.
\newblock URL \url{https://www.aclweb.org/anthology/K16-1028}.

\bibitem[Paice(1990)]{paice1990constructing}
Chris~D Paice.
\newblock Constructing literature abstracts by computer: techniques and
  prospects.
\newblock \emph{Information Processing \& Management}, 26\penalty0
  (1):\penalty0 171--186, 1990.

\bibitem[Papineni et~al.(2002)Papineni, Roukos, Ward, and Zhu]{Bleu}
Kishore Papineni, Salim Roukos, Todd Ward, and Wei-Jing Zhu.
\newblock {B}leu: a method for automatic evaluation of machine translation.
\newblock In \emph{Proceedings of the 40th Annual Meeting of the Association
  for Computational Linguistics}, pp.\  311--318, Philadelphia, Pennsylvania,
  USA, July 2002. Association for Computational Linguistics.
\newblock \doi{10.3115/1073083.1073135}.
\newblock URL \url{https://www.aclweb.org/anthology/P02-1040}.

\bibitem[Ranzato et~al.(2016)Ranzato, Chopra, Auli, and Zaremba]{ExposureBias}
Marc'Aurelio Ranzato, Sumit Chopra, Michael Auli, and Wojciech Zaremba.
\newblock Sequence level training with recurrent neural networks.
\newblock In \emph{4th International Conference on Learning Representations,
  {ICLR} 2016, San Juan, Puerto Rico, May 2-4, 2016, Conference Track
  Proceedings}, 2016.
\newblock URL \url{http://arxiv.org/abs/1511.06732}.

\bibitem[See et~al.(2017)See, Liu, and Manning]{get2point}
Abigail See, Peter Liu, and Christoper Manning.
\newblock Get to the point: Summarization with pointer-generator networks.
\newblock pp.\  1073--1083, 01 2017.
\newblock \doi{10.18653/v1/P17-1099}.

\end{thebibliography}
\bibliographystyle{iclr2020_conference}
\appendix

\end{document}